# A Data Mining Approach to the Diagnosis of Tuberculosis by Cascading Clustering and Classification

Asha.T, S. Natarajan, and K.N.B. Murthy

**Abstract**— In this paper, a methodology for the automated detection and classification of Tuberculosis(TB) is presented. Tuberculosis is a disease caused by mycobacterium which spreads through the air and attacks low immune bodies easily. Our methodology is based on clustering and classification that classifies TB into two categories, Pulmonary Tuberculosis(PTB) and retroviral PTB(RPTB) that is those with Human Immunodeficiency Virus (HIV) infection. Initially K-means clustering is used to group the TB data into two clusters and assigns classes to clusters. Subsequently multiple different classification algorithms are trained on the result set to build the final classifier model based on K-fold cross validation method. This methodology is evaluated using 700 raw TB data obtained from a city hospital. The best obtained accuracy was 98.7% from support vector machine (SVM) compared to other classifiers. The proposed approach helps doctors in their diagnosis decisions and also in their treatment planning procedures for different categories.

—————————— ◆ ——————————

## 1 INTRODUCTION

Tuberculosis is a common and often deadly infectious disease caused by mycobacterium; in humans it is mainly *Mycobacterium tuberculosis*. It is a great problem for most developing countries because of the low diagnosis and treatment opportunities. Tuberculosis has the highest mortality level among the diseases caused by a single type of microorganism. Thus, tuberculosis is a great health concern all over the world, and in India as well [wikipedia.org].

Data mining has been applied with success in different fields of human endeavour, including marketing, banking, customer relationship management, engineering and various areas of science. However, its application to the analysis of medical data has been relatively limited. Thus, there is a growing pressure for intelligent data analysis such as data mining to facilitate the extraction of knowledge to support clinical specialists in making decisions. Medical datasets have reached enormous capacities. This data may contain valuable information that awaits extraction. The knowledge may be encapsulated in various patterns and regularities that may be hidden in the data. Such knowledge may prove to be priceless in future medical decision-making. Data analysis underlies many computing applications, either in a design phase or as part of their on-line operations. Data analysis procedures can be dichotomized as either exploratory or confirmatory, based on the availability of appropriate models for the data source, but a key element in both types of procedures (whether for hypothesis formation or decision-making) is the grouping, or classification of measurements based on either goodness-of-fit to a postulated model, or natural groupings (clustering) revealed through analysis.

Clustering is the unsupervised classification of patterns (observations, data items, or feature vectors) into groups (clusters). The clustering problem has been addressed in many contexts and by researchers in many disciplines; this reflects its broad appeal and usefulness as one of the steps in exploratory data analysis. However, clustering is a difficult problem combinatorially, and differences in assumptions and contexts in different communities has made the transfer of useful generic concepts and methodologies slow to occur.

Data classification process using knowledge obtained from known historical data has been one of the most intensively studied subjects in statistics, decision science and computer science. Data mining techniques have been applied to medical services in several areas, including prediction of effectiveness of surgical procedures, medical tests, medication, and the discovery of relationships among clinical and diagnosis data. In order to help the clinicians in diagnosing the type of disease computerized data mining and decision support tools are used which are able to help clinicians to process a huge amount of data available from solving previous cases and suggest the probable diagnosis based on the values of several important attributes. There have been numerous comparisons of the different classification and prediction

————————————————
- *Asha.T is with the Dept. of Information Science & Engg., Bangalore Institute of Technology, Bangalore-560004, Karnataka, India. E-mail: asha.masthi @ gmail.com.*
- *S. Natarajan is with Dept. of Information Science & Engg., PES Institute of Technology, Bangalore-560085, Karnataka ,India. E-mail: natarajan @pes.edu.*
- *K N B Murthy is principal & Director, PES Institute of Technology, Bangalore-560085, Karnataka , India. E-mail:principal@pes.edu.*

methods, and the matter remains a research topic. No single method has been found to be superior over all others for all data sets.

It is important to understand the difference between clustering (unsupervised classification) and classification (supervised classification). In supervised classification, we are provided with a collection of *labelled* (preclassified) patterns; the problem is to label a newly encountered, yet unlabeled, pattern. Typically, the given labelled (*training*) patterns are used to learn the descriptions of classes which in turn are used to label a new pattern. In the case of clustering, the problem is to group a given collection of unlabeled patterns into meaningful clusters. In a sense, labels are associated with clusters also, but these category labels are *data driven*; that is, they are obtained solely from the data. Clustering is useful in several exploratory pattern-analysis, grouping, decision-making, and machine-learning situations, including data mining, document retrieval, image segmentation, and pattern classification. However, in many such problems, there is little prior information.

In this paper, we introduce a combined approach in the detection of Tuberculosis by cascading machine learning algorithms. K-means clustering algorithm with different classification algorithms such as Naïve Bayes, C4.5 decision trees, SVM, Adaboost and Random Forest trees etc. are combined to improve the classification accuracy of TB. In the first stage, k-Means clustering is performed on training instances to obtain k disjoint clusters. Each k-Means cluster represents a region of similar instances, "similar" in terms of Euclidean distances between the instances and their cluster centroids. We choose k-Means clustering because: 1) it is a data-driven method with relatively few assumptions on the distributions of the underlying data and 2) the greedy search strategy of k-Means guarantees at least a local minimum of the criterion function, thereby accelerating the convergence of clusters on large data sets. In the second stage, the k-Means method is cascaded with the classification algorithms to learn the classification model using the instances in each k-Means cluster.

## 2 RELATED WORK

There has been few works done on TB using Artificial neural network(ANN) and more research work has been carried out on hybrid prediction models.

Orhan Er. And Temuritus[1,2] present a study on tuberculosis diagnosis, carried out with the help of MultiLayer Neural Networks (MLNNs). For this purpose, an MLNN with one and two hidden layers and a genetic algorithm for training algorithm has been used. Data mining approach was adopted to classify genotype of mycobacterium tuberculosis using c4.5 algorithm[3]. Our proposed work is on categorical and numerical attributes of TB data with data mining technologies. Shekhar R. Gaddam et.al.[4] present "K-Means+ID3," a method to cascade k-Means clustering and the ID3 decision tree learning methods for classifying anomalous and normal activities in a computer network, an active electronic circuit, and a mechanical mass-beam system.Chin-Yuan Fan et al., propose a hybrid model [5]by integrating a case-based data clustering method and a fuzzy decision tree for medical data classification on liver disorder and breast cancer datasets. Jian kang[6] and his team propose a novel and abstract method for describing DDoS attacks with characteristic tree, three-tuple, and introduces an original, formalized taxonomy based on similarity and Hierarchical Clustering method. Yi-Hsin Yu et. al. [7] attempt to develop an EEG based classification system to automatically classify subject's Motion Sickness level and find the suitable EEG features via common feature extraction, selection and classifiers technologies in this study. Themis P. Exarchos et.al. propose a methodology[8] for the automated detection and classification of transient events in electroencephalographic (EEG) recordings. It is based on association rule mining and classifies transient events into four categories.Pascal Boilot[9] and his team report on the use of the Cyranose 320 for the detection of bacteria causing eye infections using pure laboratory cultures and the screening of bacteria associated with ENT infections using actual hospital samples. Bong-Horng chu and his team[10] propose a hybridized architecture to deal with customer retention problems.

## 3 DATA SOURCE

The medical dataset we are classifying includes 700 real records of patients suffering from TB obtained from a state hospital. The entire dataset is put in one file having many records. Each record corresponds to most relevant information of one patient. Initial queries by doctor as symptoms and some required test details of patients have been considered as main attributes. Totally there are 11 attributes(symptoms) and one class attribute. The symptoms of each patient such as age, chroniccough(weeks), loss of weight, intermittent fever(days), night sweats, Sputum, Bloodcough, chestpain, HIV, radiographic findings, wheezing and class are considered as attributes.

Table 1 shows names of 12 attributes considered along with their Data Types (DT). Type N-indicates numerical and C is categorical

## 4 PROPOSED METHOD

Figure 1 depicts the proposed hybrid model which is a combination of k-means and other classification algorithms. In the first stage the raw data collected from hospital is cleaned by filling in the missing values as null since it was not available. Second stage groups similar data into two clusters using K-means. In the third stage result set is then classified using SVM, Naïve Bayes, C4.5 Decision Tree, K-NN, Bagging, AdaBoost and RandomForest into two categories as PTB and RPTB. Their performance is evaluated using Precision, Recall, kappa statistics, Accuracy and other statistical measures.

**Table 1**
**List of Attributes and their Datatypes**

| No | Name | DT |
|----|------|----|
| 1 | Age | N |
| 2 | chroniccough(weeks) | N |
| 3 | weightloss | C |
| 4 | intermittentfever(days) | N |
| 5 | nightsweats | C |
| 6 | Bloodcough | C |
| 7 | chestpain | C |
| 8 | HIV | C |
| 9 | Radiographicfindings | C |
| 10 | Sputum | C |
| 11 | wheezing | C |
| 12 | class | C |

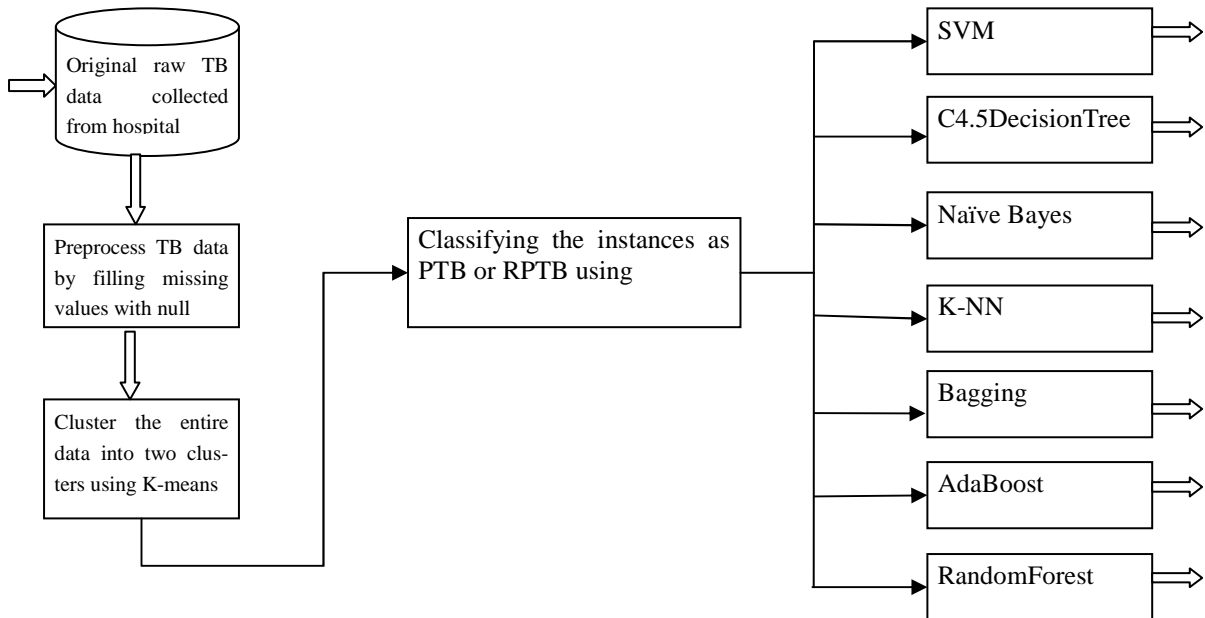

Fig. 1 proposed combined approach to cluster-classification

## 5 ALGORITHMS

### 5.1 K-Means Clustering

K-means clustering is an algorithm[20] to classify or to group objects based on attributes into K number of group. K is a positive integer number. The grouping is done by minimizing the sum of squares of distances between data and the corresponding cluster centroid. It can be viewed as a greedy algorithm for partitioning the n samples into k clusters so as to minimize the sum of the squared distances to the cluster centres. It does have some weaknesses: The way to initialize the means was not specified. One popular way to start is to randomly choose k of the samples. The basic step of direct k-means clustering is simple. In the beginning we determine number of cluster k and we assume the centroid or centre of these clusters. Let the K prototypes $(w_1..........w_k)$ be initialized to one of the input patterns $(i_1..........i_n)$. Where $w_j \in i_l$, $j \in 1,.........,k$, $l \in 1,.......nC_j$ is the $j^{th}$ cluster whose value is a disjoint subset of input patterns. The quality of the clustering is determined by the following error function:

$$E = \sum_{j=1}^{k} \sum_{i_l \in c_j} |i_l - w_j|^2$$

The appropriate choice of *k* is problem and domain dependent and generally a user tries several values of *k*. Assuming that there are *n* patterns, each of dimension *d*, the computational cost of a direct k-means algorithm per iteration can be decomposed into three parts.

- The time required for the first *for* loop in the algorithm is $O(nkd)$.
- The time required for calculating the centroids is $0(nd)$.
- The time required for calculating the error function is $O(nd)$.

### 5.2 C4.5 Decision Tree

Perhaps C4.5 algorithm which was developed by Quinlan is the most popular tree classifier[21]. It is a decision support tool that uses a tree-like graph or model of decisions and their possible consequences, including chance event outcomes, resource costs, and utility. Weka classifier package has its own version of C4.5 known as J48. J48 is an optimized implementation of C4.5 rev. 8.

### 5.3 K-Nearest Neighbor(K-NN)

The *k*-nearest neighbors algorithm (*k*-NN) is a method for[22] classifying objects based on closest training examples in the feature space. *k*-NN is a type of instance-based learning., or lazy learning where the function is only approximated locally and all computation is deferred until classification. Here an object is classified by a majority vote of its neighbors, with the object being assigned to the class most common amongst its *k* nearest neighbors (*k* is a positive, typically small).

### 5.4 Naïve Bayesian Classifier

It is Bayes classifier which is a simple probabilistic classifier based on applying Baye's theorem(from Bayesian statistics) with strong (naive) independence[23] assumptions. In probability theory Bayes' theorem shows how one conditional probability (such as the probability of a hypothesis given observed evidence) depends on its inverse (in this case, the probability of that evidence given the hypothesis). In more technical terms, the theorem expresses the posterior probability (i.e. after evidence E is observed) of a hypothesis H in terms of the prior probabilities of H and E, and the probability of E given H. It implies that evidence has a stronger confirming effect if it was more unlikely before being observed.

### 5.5 Support Vector Machine

The original SVM algorithm was invented by Vladimir Vapnik. The standard SVM takes a set of input data, and predicts, for each given input, which of two possible classes the input is a member of, which makes the SVM a non-probabilistic binary linear classifier.

A support vector machine constructs a hyperplane or set of hyperplanes in a high or infinite dimensional space, which can be used for classification, regression or other tasks. Intuitively, a good separation is achieved by the hyperplane that has the largest distance to the nearest training data points of any class (so-called functional margin), since in general the larger the margin the lower the generalization error of the classifier.

### 5.6 Bagging

Bagging (Bootstrap aggregating) was proposed by Leo Breiman in 1994 to improve the classification by combining classifications of randomly generated training sets. The concept of bagging[24] (voting for classification, averaging for regression-type problems with continuous dependent variables of interest) applies to the area of predictive data mining to combine the predicted classifications (prediction) from multiple models, or from the same type of model for different learning data. It is a technique generating multiple training sets by sampling with replacement from the available training data and assigns vote for each classification.

### 5.7 Random Forest

The algorithm for inducing a random forest was developed by leo-braiman[25]. The term came from random decision forests that was first proposed by Tin Kam Ho of Bell Labs in 1995. It is an ensemble classifier that consists of many decision trees and outputs the class that is the mode of the class's output by individual trees. It is a popular algorithm which builds a randomized decision tree in each iteration of the bagging algorithm and often produces excellent predictors.

### 5.8 Adaboost

AdaBoost is an algorithm for constructing a "strong" classifier as linear combination of "simple" "weak" classifier. Instead of resampling, Each training sample uses a weight to determine the probability of being selected for a training set. Final classification is based on weighted vote of weak classifiers. AdaBoost is sensitive to noisy data and outliers. However in some problems it can be less susceptible to the overfitting problem than most learning algorithms.

## 6 PERFORMANCE MEASURES

Some measure of evaluating performance has to be introduced. One common measure in the literature (Chawla, Bowyer, Hall & Kegelmeyer, 2002) is accuracy defined as correct classified instances divided by the total number of instances. A single prediction has the four different possible outcomes shown from confusion matrix in Table 2. The true positives (TP) and true negatives (TN) are correct classifications. A false positive (FP) occurs when the outcome is incorrectly predicted as yes (or positive) when it is actually no (negative). A false negative (FN) occurs when the outcome is incorrectly predicted as no when it is actually yes. Various measures used in this study are:

Accuracy = (TP + TN) / (TP + TN + FP + FN)
Precision = TP / (TP + FP)
Recall / Sensitivity = TP / (TP + FN)

TABLE 2
Confusion Matrix

|  |  | Predicted Label | |
|---|---|---|---|
|  |  | *Positive* | *Negative* |
| **Known Label** | *Positive* | True Positive (TP) | False Negative (FN) |
|  | *Negative* | False Positive (FP) | True Negative (TN) |

*k-Fold cross-validation*: In order to have a good measure performance of the classifier, k-fold cross-validation method has been used (Delen et al.,2005).The classification algorithm is trained and tested k time.In the most elementary form, cross validation consists of dividing the data into k subgroups. Each subgroup is tested via classification rule constructed from the remaining (k - 1) groups. Thus the k different test results are obtained for each train–test configuration. The average result gives the test accuracy of the algorithm. We used 10 fold cross-validations in our approach. It reduces the bias associated with random sampling method.

*Kappa statistics*: The kappa parameter measures pair wise agreement between two different observers, corrected for an expected chance agreement (Thora, Ebba, Helgi & Sven,2008). For example if the value is 1, then it means that there is a complete agreement between the classifier and real world value. Kappa value can be calculated from following formula
K = [P(A) - P(E)] / [1-P(E)]
where P(A) is the percentage of agreement between the classifier and underlying truth calculated. P(E) is the chance of agreement calculated.

## 7 EXPERIMENTAL RESULTS

For the implementation we have used Waikato Environment for Knowledge Analysis (WEKA) toolkit to analyze the performance gain that can be obtained by using various classifiers (Witten & Frank, 2000). WEKA[26] consists of number of standard machine learning methods that can be applied to obtain useful knowledge from databases which are too large to be analyzed by hand. Machine learning algorithms differ from statistical methods in the way that it uses only useful features from the dataset for analysis based on learning techniques.Table 3 displays the comparison of different measures such as Mean absolute error, Relative absolute error with kappa statistics whereas Table 4 lists accuracy, F-measure and Incorrectly classified instances of multiple classifiers mentioned above.

Table 3 Experimental Results of various statistical measures.

| Clusters | Classifiers | Class category | Precision | Recall | Mean absolute Error | Relative absolute Error | Kappa Statistics |
|---|---|---|---|---|---|---|---|
| Cluster 0 | SVM | PTB | 98.5% | 99.3% | 0.0129 | 2.6387% | 0.9736 |
| Cluster 1 |  | RPTB | 99% | 98% |  |  |  |
| Cluster 0 | C4.5DecisionTree | PTB | 91.9% | 95.3% | 0.1323 | 27.1585% | 0.8435 |
| Cluster 1 |  | RPTB | 93.2% | 88.4% |  |  |  |
| Cluster 0 | NaiveBayes | PTB | 93% | 91.9% | 0.1388 | 28.49% | 0.8216 |
| Cluster 1 |  | RPTB | 89% | 90% |  |  |  |
| Cluster 0 | K-NN | PTB | 96.3% | 95.8% | 0.0472 | 9.6771% | 0.9063 |
| Cluster 1 |  | RPTB | 94.3% | 94.9% |  |  |  |
| Cluster 0 | Bagging | PTB | 98.1% | 99.3% | 0.0336 | 6.8966% | 0.9677 |
| Cluster 1 |  | RPTB | 99% | 97.3% |  |  |  |
| Cluster 0 | AdaBoost | PTB | 96.9% | 99.3% | 0.0524 | 10.746% | 0.9529 |
| Cluster 1 |  | RPTB | 98.9% | 95.6% |  |  |  |
| Cluster 0 | RandomForest | PTB | 98.5% | 98.5% | 0.0932 | 19.1267% | 0.9648 |
| Cluster 1 |  | RPTB | 98% | 98% |  |  |  |

Table 4 Comparison of Accuracy with other measures on different classifiers.

| Classifiers | Accuracy | F-measure | Incorrect classification |
|---|---|---|---|
| ANN(Existing Result in Ref.[1]) | 93% | - | - |
| SVM | 98.7% | 0.987 | 1.2857% |
| C4.5DecisionTree | 92.4% | 0.924 | 7.57% |
| NaiveBayes | 91.3% | 0.913 | 8.7% |
| K-NN | 95.4% | 0.954 | 4.5714% |
| Bagging | 98.4% | 0.984 | 1.5714% |
| AdaBoost | 97.7% | 0.977 | 2.2857% |
| RandomForest | 98.3% | 0.983 | 1.7143% |

It can be seen from tables that SVM has highest accuracy followed by Bagging and RandomForest Trees compared to other classifiers.

A Graph showing in detail the comparison of accuracy, True positive Rate(TPR), ROC area has been shown in figure 2 and figure 3 respectively.

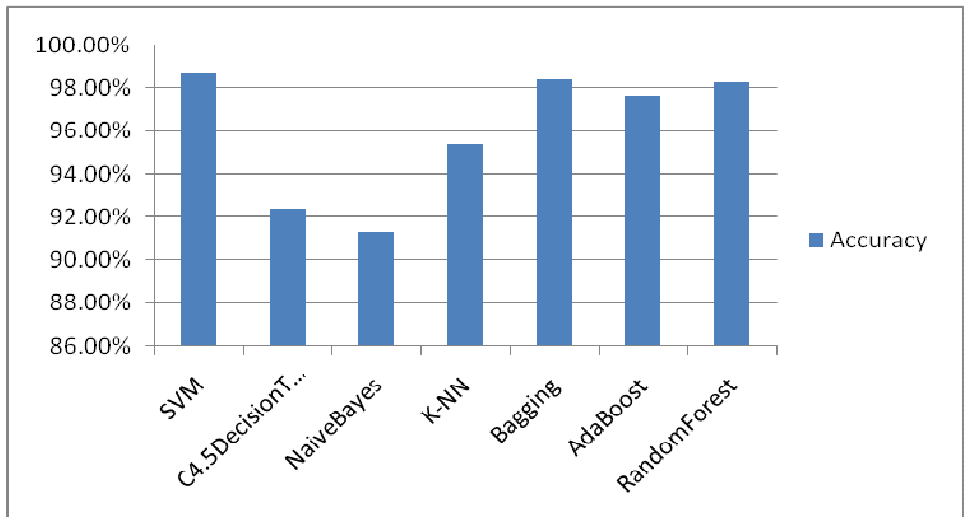

Fig.2 Performance Comparison of all the classifiers.

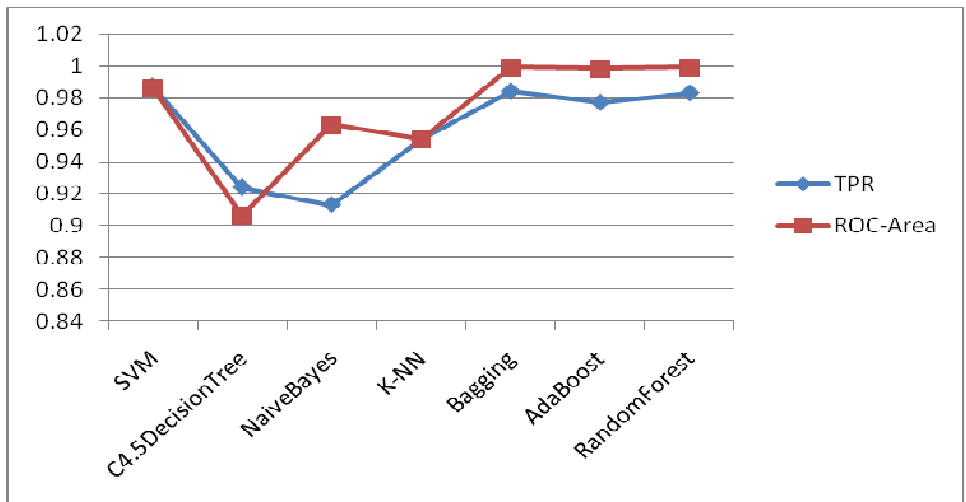

Fig.3 Comparison of True Positive rate and F-measure.

## CONCLUSION

Tuberculosis is an important health concern as it is also associated with AIDS. Retrospective studies of tuberculosis suggest that active tuberculosis accelerates the progression of HIV infection. In this paper, we propose an efficient hybrid model for the prediction of tuberculosis. K-means clustering is combined with various different classifiers to improve the accuracy in the prediction of TB. This approach not only helps doctors in diagnosis but also to consider various other features involved within each class in planning their treatments. Compared to existing NN classifiers and NN with GA, our model produces an accuracy of 98.7% with SVM.

## ACKNOWLEDGMENT

Our thanks to KIMS Hospital, Bangalore for providing the valuable real Tuberculosis data and principal Dr. Sudharshan for giving permission to collect data from the Hospital.